\documentclass{article}

\usepackage[preprint]{corl_2026}

\PassOptionsToPackage{sort}{natbib}

\usepackage[utf8]{inputenc} 
\usepackage[T1]{fontenc}    
\usepackage{url}            
\usepackage{booktabs}       
\usepackage{amsfonts}
\usepackage{nicefrac}
\usepackage{microtype}     
\usepackage{xcolor}
\usepackage{graphicx}

\usepackage{epsfig}
\usepackage{amsmath}
\usepackage{amssymb}
\usepackage{wrapfig}
\usepackage{caption}
\usepackage[font=tiny,labelfont=bf]{subcaption}
\usepackage{pifont}
\usepackage{enumitem}
\usepackage{tabularx}
\usepackage{multirow}
\usepackage{hyperref}

\setitemize{noitemsep,topsep=0pt,parsep=0pt,partopsep=0pt}

\definecolor{citecolor}{HTML}{0071BC}
\definecolor{linkcolor}{HTML}{ED1C24}
\hypersetup{citecolor=citecolor,linkcolor=linkcolor}

\def\VspaceS{\vspace{-0.00cm}}
\def\VspaceL{\vspace{-0.00cm}}

\title{From Fixed to Free Cameras: Calibration-Free View-Robust Vision-Language-Action Model}

\author{
Wenhao Li\textsuperscript{1},
Xueying Jiang\textsuperscript{1},
Quanhao Qian\textsuperscript{2,3},
Deli Zhao\textsuperscript{2,3}, \\
\textbf{Shijian Lu}\textsuperscript{1}\textsuperscript{\ding{41}}, 
\textbf{Gongjie Zhang}\textsuperscript{4}\textsuperscript{\ding{41}},
\textbf{Ran Xu}\textsuperscript{2,3}\textsuperscript{\ding{41}}
\\
\textsuperscript{1}Nanyang Technological University \\
\textsuperscript{2}DAMO Academy, Alibaba Group \quad
\textsuperscript{3}HuPan Lab \quad
\textsuperscript{4}Alibaba Group
}

\begin{document}
\maketitle

\begin{abstract}
Real-world robot deployment rarely maintains the training-stage camera setup, where cameras often experience repositioning or remounting depending on actual scenarios. Existing view-robust Vision-Language-Action (VLA) policies tolerate such camera variations only when the camera extrinsics are explicitly provided, making them fragile and hard to use especially when view robustness is critical. We argue that the policy should not be told where the camera is, but rather figure it out by itself. To this end, we introduce Camera-Centric VLA (CamVLA), a new VLA model that decouples manipulation controls from camera geometry by predicting (i) a camera-centric end-effector action expressed in the local camera frame, and (ii) a 6-DoF hand-eye matrix relating cameras to the robot base. A deterministic geometric transformation composes the two predictions into a robot base-frame action. This disentangles \textit{how I should move} in pose-independent camera-centric action generation from \textit{where I am looking from} in camera-perspective geometric grounding. The resulting policy is calibration-free, depth-free, and single-view, requiring only a single monocular RGB image as the visual observation and task instruction at deployment. Evaluations in both simulation and real-world robot data show that CamVLA consistently improves success rates across diverse unseen viewpoints. Project page: \url{https://alibaba-damo-academy.github.io/CamVLA/}. 
\end{abstract}


\section{Introduction}

Vision-Language-Action (VLA) models~\cite{rt1, openvlaoft, pi07, gr00t} have rapidly progressed toward generalist robot policies, leveraging internet-scale vision-language data and diverse robotic demonstrations to ground broad semantic knowledge into directly executable manipulation. 
Yet despite their semantic competence, state-of-the-art VLAs exhibit a sharp and unexpected brittleness to camera viewpoint shifts. 

\begin{wrapfigure}{r}{0.5\textwidth}
  \centering
  \vspace{-18pt}
  \includegraphics[width=0.48\textwidth]{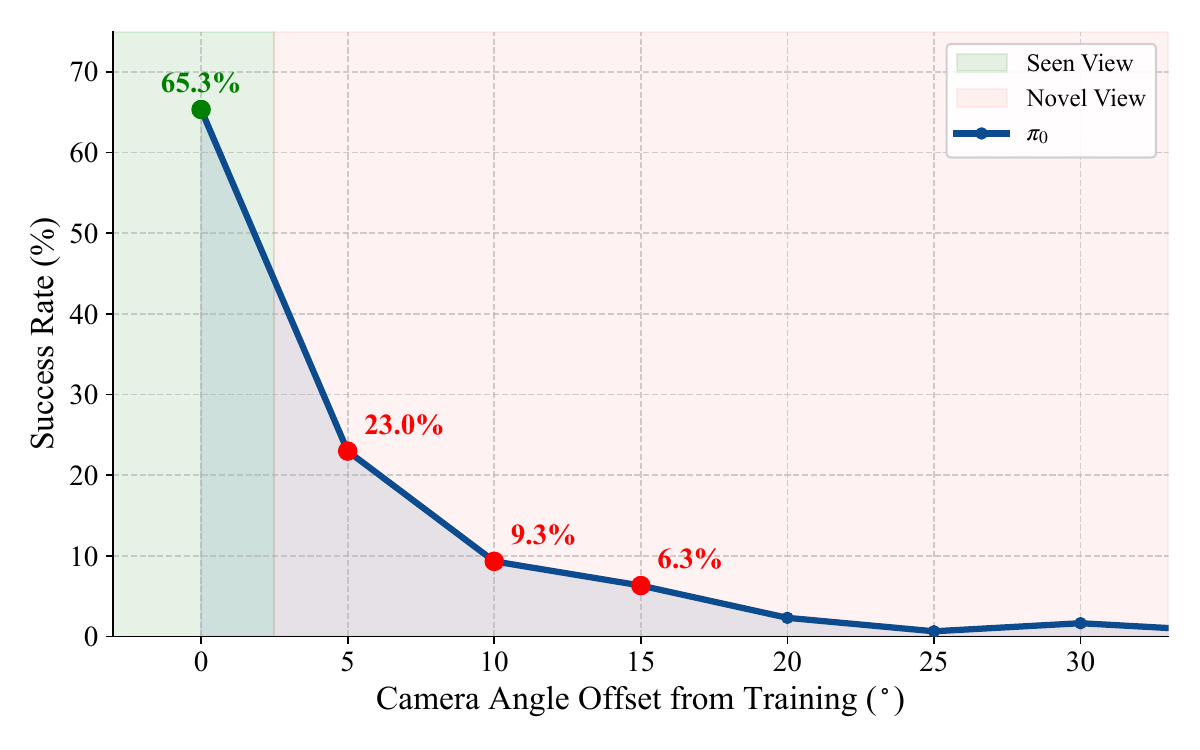}
  \vspace{-6pt}
  \caption{\textbf{The Viewpoint Trap in VLAs.}  
  Conventional VLA (\textit{e.g.}, $\pi_{0}$) trained on a single viewpoint exhibits extreme spatial brittleness, where a mere 15$^\circ$ camera shift drops success rates to 6.3\%.}
  \label{fig:teaser}
  \vspace{-10pt}
\end{wrapfigure}

As illustrated in Figure~\ref{fig:teaser}, $\pi_{0}$~\cite{pi0} trained on a single canonical perspective achieves a $\sim$65.3\% success rate under its training view on RLBench~\cite{james2020rlbench}, yet collapses to a mere 6.3\% under a 15$^\circ$ camera rotation. 
This failure persists even when the scene remains fully observable and the semantic goal is unchanged. 
Although large-scale multi-view training could mitigate this, acquiring such data is prohibitively expensive and hard to scale. 
In practice, real-world robot deployment rarely matches the controlled camera setup of training time: sensors get bumped during operation, mounted on different platforms, hand held by operators, or affixed to mobile bases whose pose drifts continuously. 
Consequently, without inherent view robustness, VLAs remain tethered to static laboratory setups, failing to generalize to the dynamic and unconstrained configurations of real-world deployment. 

This brittleness has a structural origin. 
Standard VLAs \cite{openvla, pi0, gr00t} rigidly predict actions in the robot base frame from camera-perspective visual observations. 
However, this base-frame parameterization misaligns action outputs with camera-frame inputs, requiring the network to implicitly resolve the spatial transformation from the camera to the robot base (hand-eye transformation). 
Without explicit geometric constraints, this hand-eye transformation remains a hidden variable, forcing the policy to memorize coordinate mappings by coupling manipulation control with camera geometry, which easily collapses under minor viewpoint shifts. 
Recent works converge on a single recipe to fix this by \textit{telling the policy where the camera is}. 
For instance, OC-VLA~\cite{zhang2026grounding} bypasses this frame-of-reference gap by re-expressing targets in calibrated camera coordinates; 
Jiang et al.~\cite{jiang2025you} rely on ray embeddings derived from known camera parameters; 4D-VLA~\cite{zhang20254d} back-projects pixels into the robot frame using known intrinsics and extrinsics; and AnyCamVLA~\cite{heo2026anycamvla} synthesizes canonical views given both source and target camera poses. 
Despite their architectural diversity, all of these methods require \textit{known and accurate camera extrinsics at deployment}, which is precisely the assumption that breaks under hand-held, drifting, or remounted cameras. 
Existing view-robust VLAs are therefore most fragile in the deployment regimes that motivate view robustness in the first place. 

In this work, we argue that the policy should not be \textit{told} where the camera is, but rather should \textit{figure it out by itself}. 
This self-localizing capability finds a natural analogue in human cognition: human vision-guided manipulation operates natively in an egocentric reference frame, while a complementary allocentric system maintains an implicit understanding of head pose relative to the torso~\cite{goodale1992separate, crawford2004spatial}. 
Inspired by this biological factorization, we decouple the VLA policy into two distinct subproblems, both inferable from a single monocular RGB image:
\textit{(i) Camera-Centric Action Generation} answers \textit{``How should I move?''} by predicting end-effector actions natively in the local camera coordinate frame, which aligns directly with visual observations, making the mapping inherently independent of the external camera pose. 
\textit{(ii) Camera-Perspective Geometric Grounding} answers \textit{``Where am I looking from?''} by regressing the 6-DoF hand-eye matrix that relates the observing camera to the robot base, explicitly modeling the relative spatial geometry and isolating viewpoint-dependent variation within a single relative pose. 

\begin{figure*}[t]
\centering
\includegraphics[width=\textwidth]{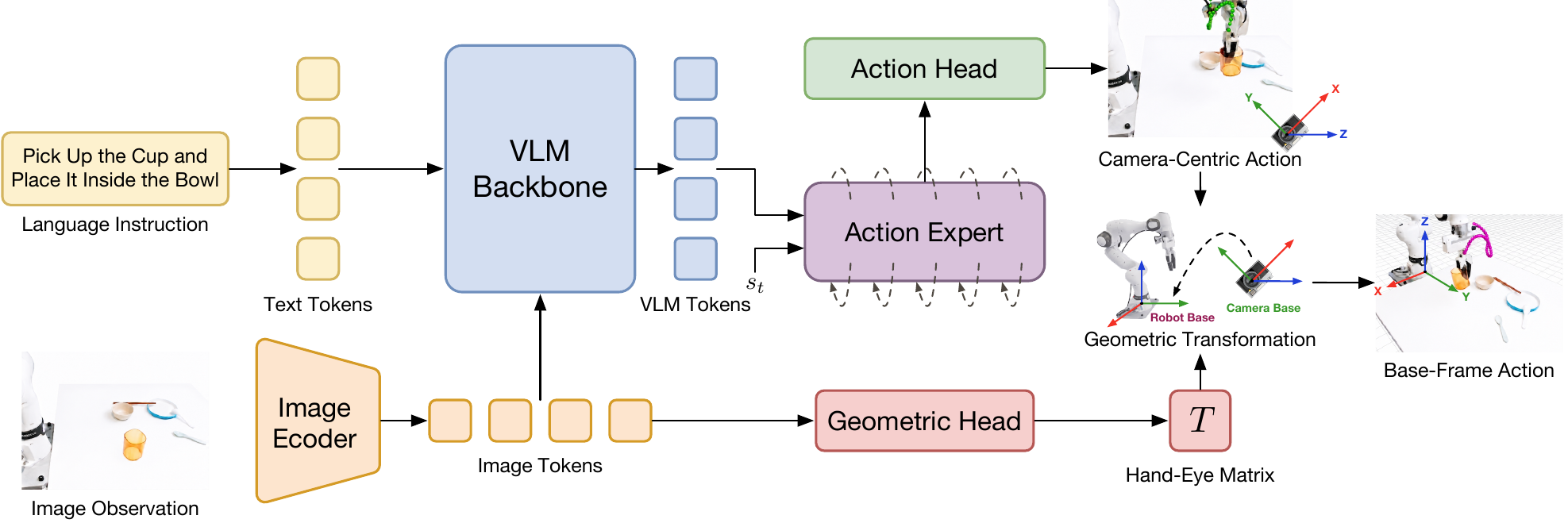}
\VspaceS
\caption{\textbf{Overview of the CamVLA Architecture.}
Our CamVLA predicts the local camera-centric action and the 6-DoF hand-eye pose in parallel, which are then combined via a deterministic geometric transformation to execute the base-frame action.}
\VspaceL
\label{fig:arch}
\end{figure*}

We instantiate this factorization as CamVLA (see Figure~\ref{fig:arch}), a calibration-free, camera-centric VLA model. 
From the extracted visual representations, two specialized heads predict the camera-centric action and the hand-eye matrix in parallel, and a deterministic geometric transformation composes them into a robot base-frame action. 
Parameterizing actions natively in the camera frame aligns them directly with visual observations, avoiding the geometric entanglement of base-frame policies that map conflicting visual flows to base coordinates. 
By absorbing all viewpoint variability into a single learned 6-DoF pose, CamVLA presents a simpler regression task than memorizing view-dependent coordinate mappings. 
Crucially, it requires no external camera information at deployment (e.g., calibrated extrinsics, depth sensors, or view synthesis), relying exclusively on the same monocular RGB image and language command that standard VLA already receives. 
By replacing the assumption of \textit{given} geometry with the discipline of \textit{learned} geometry, CamVLA closes the deployment gap left open by prior view-robust methods, enabling calibration-free manipulation under uncalibrated viewpoint shifts on real hardware. 

Our contributions can be summarized as threefold. 
\textit{First}, we identify that existing view-robust VLA approaches share a common, deployment-fragile assumption of known camera extrinsics, and we argue for a self-localizing alternative that infers camera geometry directly from RGB. 
\textit{Second}, we propose CamVLA, a calibration-free, depth-free, single-view VLA framework that decouples camera-centric action generation from camera-perspective geometric grounding and recombines them via a deterministic geometric transformation. 
\textit{Third}, we comprehensively evaluate CamVLA in both simulation and real-world deployment, demonstrating substantial success rate improvements over strong VLA baselines (e.g., $\pi_{0}$~\cite{pi0} and GR00T N1.7~\cite{gr00t}) across diverse unseen camera configurations. 

\section{Related Work}

\noindent \textbf{Vision-Language-Action Models.} 
The pursuit of generalist robot policies has catalyzed the development of VLAs \cite{rt2, driess2023palm, ahn2022can, openx, pi06}, which adapt VLMs \cite{liu2023visual, li2023blip, zhu2024minigpt, Qwen2.5-VL} by predicting continuous robot actions via large-scale transformers. 
Octo \cite{octo} utilizes a Transformer backbone to handle multi-modal observations and a diffusion head \cite{chi2025diffusion} for robust action generation. 
Recent flow-matching architectures such as $\pi_0$ \cite{pi0} and $\pi_{0.5}$ \cite{pi05} have advanced high-frequency continuous control, open-world generalization, and learning from real-world experience. 
GR00T N1 \cite{gr00t} employs a dual-system architecture to coordinate high-level reasoning with low-level motor commands for humanoid control. 
Fast-in-Slow \cite{chen2025fast} further explores the unification of fast manipulation within a slow reasoning system to improve real-time responsiveness. 
While these models possess strong semantic knowledge, their generalization capabilities remain limited under novel camera viewpoints. 

\noindent \textbf{Viewpoint Generalization in Robotics.} 
Achieving viewpoint robustness remains a critical bottleneck for robot policies, whose performance can degrade sharply when the deployment camera pose differs from the training setup \cite{li2025vla}. 
One line of work introduces explicit 3D structure to improve geometric consistency. 
For example, Perceiver-Actor \cite{shridhar2023perceiver} builds voxelized RGB-D scene representations, and recent VLA models incorporate 3D features, depth/point-cloud priors, or spatiotemporal 3D representations \cite{zhen20243d, jia2024lift3d, zhang2025spatial, ke20243d, deng2025stereovla}. 
However, these approaches often introduce computational overhead and depend on depth sensing, multi-view observations, or calibrated camera geometry. 
Alternative approaches improve robustness via novel-view synthesis or image-level augmentation \cite{tian2024view, coholich2026sim2real, xu2025egodemogen}, viewpoint selection \cite{vasudevan2025strategic}, and view-invariant representations or latent actions \cite{pang2025learning, jeong2026learning, luo2025viewpoint, li2026manivid, sun2026view, abouzeid2025geoaware}. 
Several concurrent works mitigate viewpoint shifts via camera-aware policy conditioning, including grounding actions in the camera space~\cite{zhang2026grounding}, conditioning on camera parameters via ray embeddings~\cite{jiang2025you}, back-projecting pixels into the robot base frame using known camera parameters~\cite{zhang20254d}, or synthesizing canonical views based on relative camera poses~\cite{heo2026anycamvla}. 
Unlike these camera-aware methods that all heavily rely on known and accurate camera intrinsics or extrinsics at deployment, CamVLA self-predicts the hand-eye matrix directly from monocular RGB, achieving view-robust manipulation without external calibration, depth sensing, novel-view synthesis, or 3D reconstruction. 

\section{Methodology}
As shown in Figure~\ref{fig:arch}, CamVLA achieves viewpoint robustness by decoupling the policy into two parallel components: (i) an Action Head that predicts end-effector actions natively in the camera coordinate frame, and (ii) a Geometric Head that regresses the 6-DoF hand-eye matrix to estimate the camera pose relative to the robot base. 
A deterministic transformation then composes these two outputs into a base-frame action for execution. 
This decoupling design factorizes the VLA policy into pose-independent action generation and viewpoint-dependent hand-eye grounding, isolating viewpoint variations from the core visual-action mapping. 
In the following subsections, we review the standard VLA formulation and detail our decoupling pipeline. 

\subsection{Standard VLA Action Formulation}
At each timestep $t$, the VLA model receives a visual observation $I_t \in \mathbb{R}^{H \times W \times 3}$, a proprioceptive state $s_t = [p_{b, t}, r_{b, t}]$, and a natural language goal $L$. Here, $p_{b, t} \in \mathbb{R}^3$ and $r_{b, t} \in \mathbb{R}^3$ denote the base-frame position and axis-angle orientation of the robot end-effector, respectively. 
Standard VLAs \cite{openvlaoft,pi0,gr00t} predict the base-frame delta action $\Delta A_{b,t} = [\Delta p_{b,t}, \Delta r_{b,t}, g_t]$, where $\Delta p_{b,t}, \Delta r_{b,t} \in \mathbb{R}^3$ denote the translational displacement and axis-angle delta rotation, and $g_t \in [0, 1]$ is the gripper state. 
The network parameters $\theta$ are optimized with the action prediction objective $\mathcal{L}_{\text{act}}$; for regression-style policies, this can be written as $\sum_{t} \| f_\theta(I_t, s_t, L) - \Delta A_{b,t} \|_2^2$. 

\subsection{Camera-Centric Action Generation}
When $f_\theta$ is trained to output $\Delta A_{b,t}$, it learns a difficult cross-frame mapping $F: (I_t, s_t, L) \mapsto \Delta A_{b,t}$ from camera-perspective visual observations to a base-frame action. 
The bridge between these two frames is the hand-eye matrix $T_t$, a rigid transform relating the camera and robot coordinate systems~\cite{tsai1989new,shiu1987calibration}. 
Absent from the input, $T_t$ is implicitly entangled in the weights of $f_\theta$. 
While training on multi-view datasets commonly improves viewpoint generalization~\cite{qian2025gp3,cai2026beyond}, a base-frame policy forces the network to map conflicting visual flows to the same base-frame action, leading to geometric entanglement. 
Consequently, the mapping $F$ collapses once the deployment $T_t$ diverges from the training distribution. 

Rather than relying on this cross-frame mapping to absorb $T_t$ implicitly, our approach decouples the visual-to-action mapping from $T_t$ by letting the policy learn a simpler same-frame form in which both visual observations and actions are natively expressed in the local camera frame, where this mapping remains independent of how the camera is mounted. 
We define the camera-centric delta action as:
\begin{equation}
    \Delta A_{c,t} = [\Delta p_{c,t}, \Delta r_{c,t}, g_t],
\end{equation}
where $\Delta p_{c,t} \in \mathbb{R}^3$ and $\Delta r_{c,t} \in \mathbb{R}^3$ are the relative translation and axis-angle rotation expressed in the camera frame. 
Because both visual representations and actions share the same egocentric perspective, visual flows are naturally aligned with physical movement in the camera frame. 
For instance, a leftward visual translation in the image consistently corresponds to a negative displacement along the local X-axis of the camera. 
By resolving this conflict, this consistent spatial relationship prevents visual-action confusion, helping the policy generalize to unseen viewpoints. 

\subsection{Camera-Perspective Geometric Grounding}
While predicting $\Delta A_{c,t}$ ensures robust visual alignment, robotic arms fundamentally operate based on kinematic models and inverse kinematics defined in their grounded base frame~\cite{siciliano2009robotics}. 
The hand-eye transformation defines the rigid-body mapping between the observing camera and the robot base coordinate system, which is essential for bridging the visual and physical action spaces. 
Consequently, we cannot send $\Delta A_{c,t}$ directly to the low-level controller without first applying this transformation. 

To bridge this gap, our architecture incorporates a specialized auxiliary Geometric Head that regresses the 6-DoF hand-eye matrix $T_t \in SE(3)$ from visual features. 
This makes $T_t$ an explicit network output, whereas in the cross-frame baseline, it remains implicitly entangled in the weights. 
We parameterize $T_t$ with a translation vector $\tau_t \in \mathbb{R}^3$ and an axis-angle rotation vector $\omega_t \in \mathbb{R}^3$, where $\omega_t$ is converted to a rotation matrix $R_t \in SO(3)$ during geometric fusion. 
The network $f_\theta$ is therefore redefined to jointly output the camera-centric action $\Delta A_{c,t}$ from the Action Head and the hand-eye matrix $T_t$ from the Geometric Head:
\begin{equation}
    (\Delta A_{c,t}, T_t) = f_{\theta}(I_t, s_t, L).
\end{equation}

\subsection{Deterministic Geometric Transformation} 
To execute actions on the physical robot, we combine the parallel predictions from the Action Head and the Geometric Head using a deterministic geometric transformation. 
Specifically, since the relative translation $\Delta p_{c,t}$ and axis-angle rotation $\Delta r_{c,t}$ are free vectors, both transform linearly under the predicted rotation $R_t \in SO(3)$, independently of the translation vector $\tau_t$~\cite{murray2017mathematical}:
\begin{align}
    \Delta p_{b,t} &= R_t \Delta p_{c,t}, \label{eq:pos_trans} \\
    \Delta r_{b,t} &= R_t \Delta r_{c,t}. \label{eq:rot_trans}
\end{align}
The final base-frame delta action is then assembled as $\Delta A_{b,t} = [\Delta p_{b,t}, \Delta r_{b,t}, g_t]$.
Despite this mathematical independence, we still regress $\tau_t$ to enhance the geometric grounding of the visual backbone and to support potential absolute-action variants. 
Consequently, test-time drift in $\tau_t$ has zero physical impact on execution, confining viewpoint errors exclusively to $R_t$. 

\section{Experiments}
\label{sec:experiments}

\subsection{Simulation Experiments}

\begin{wrapfigure}{r}{0.5\textwidth}
  \centering
  \vspace{-10pt}
  \includegraphics[width=0.48\textwidth]{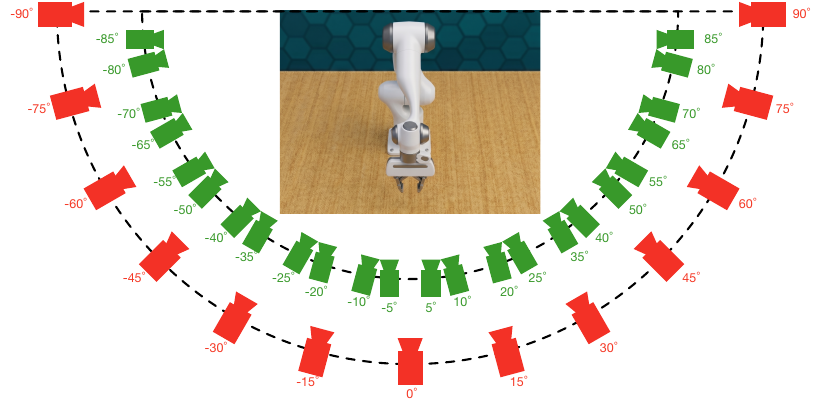}
  \vspace{-5pt}
  \caption{\textbf{Simulation camera configuration.} 
The training set (red cameras) covers discrete viewpoints, while evaluation (green cameras) is conducted on a dense set of unseen viewpoints.}
  \label{fig:sim_camera_setup}
  \vspace{-10pt}
\end{wrapfigure}
\textbf{Experimental Setup.} 
We evaluate CamVLA on the RLBench benchmark~\cite{james2020rlbench}. 
To evaluate viewpoint robustness, we utilize the front camera and rotate it around the robot base from $-90^\circ$ to $90^\circ$ at $5^\circ$ intervals to generate a diverse set of viewpoints as shown in Figure~\ref{fig:sim_camera_setup}. 
The model is trained on a discrete subset of viewpoints with $15^\circ$ intervals ($0^\circ, \pm 15^\circ, \pm 30^\circ, \dots, \pm 90^\circ$), while zero-shot generalization is evaluated on the remaining unseen viewpoints within the $5^\circ$ grid. 
We evaluate CamVLA on six representative manipulation tasks: \textit{slide block to target}, \textit{push buttons}, \textit{take umbrella out of umbrella stand}, \textit{close laptop lid}, \textit{lamp off}, and \textit{put knife on chopping board}. 
For each task and viewpoint, we collect 100 expert demonstrations for training and 50 episodes for evaluation. 
\begin{table}[t]
\centering
\scriptsize
\caption{\textbf{Zero-shot viewpoint generalization on RLBench simulation experiments.} 
Success rate (\%) comparison between VLA baselines and CamVLA across six tasks under unseen viewpoints.}
\label{tab:simulation_results}
\resizebox{\columnwidth}{!}{
\begin{tabular}{lccccccc}
\toprule
Model & Slide Block & Push Buttons & Take Umbrella & Close Laptop & Lamp Off & Put Knife & \textbf{Mean} \\
\midrule
$\pi_0$ \cite{pi0} & 18.3 & 51.5 & 32.3 & 57.0 & 29.8 & 10.0 & 33.2 \\
$\pi_0$ + CamVLA (Ours) & 44.5 & 72.3 & 39.2 & 69.0 & 58.0 & 25.3 & 51.4 \\

\midrule

GR00T N1.7 \cite{gr00t} & 27.5 & 13.5 & 41.8 & 47.7 & 28.2 & 11.5 & 28.4 \\
GR00T N1.7 + CamVLA (Ours) & 44.7 & 30.5 & 50.3 & 56.0 & 35.0 & 14.0 & 38.4 \\
\bottomrule
\end{tabular}
\VspaceL
}
\end{table}

\noindent \textbf{Zero-Shot Viewpoint Generalization.}
As shown in Table~\ref{tab:simulation_results}, on RLBench simulation experiments, we integrate our CamVLA framework with two foundational VLA architectures, $\pi_0$~\cite{pi0} and GR00T N1.7~\cite{gr00t}, evaluating zero-shot generalization across six tasks under unseen camera configurations. 
The reported success rates are averaged across all unseen viewpoints.
For $\pi_0$, integrating CamVLA improves the average success rate from 33.2\% to 51.4\% (+18.2\% absolute gain) on unseen configurations. 
Notably, on challenging tasks where the vanilla policy severely degrades under viewpoint variations, such as Slide Block and Lamp Off, CamVLA boosts the success rates from 18.3\% to 44.5\% and 29.8\% to 58.0\%, respectively. 
Similarly, applying our method to the GR00T N1.7 architecture demonstrates consistent robustness enhancements, achieving +10.0\% absolute gain (from 28.4\% to 38.4\%) on unseen configurations. 
These results confirm that CamVLA generalizes zero-shot to unseen viewpoints, consistently outperforming conventional VLA policies under viewpoint shifts. 

\subsection{Real-World Experiments}

\begin{wrapfigure}{r}{0.5\textwidth}
  \centering
  \vspace{-10pt}
  \includegraphics[width=0.48\textwidth]{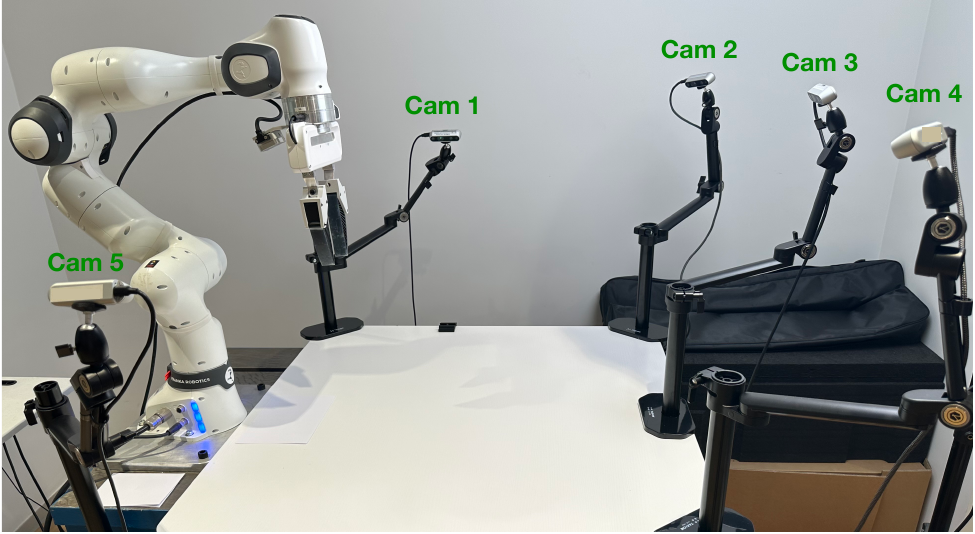}
  \vspace{-5pt}
  \caption{\textbf{Real-world experimental setup.} 
Multi-view camera configuration used to verify viewpoint robustness in physical environments.}
  \label{fig:real_camera_setup}
  \vspace{-10pt}
\end{wrapfigure}
\textbf{Experimental Setup.} 
Our real-world setup uses a Franka Research 3 robot arm with a parallel gripper. 
As shown in Figure~\ref{fig:real_camera_setup}, we use a multi-view camera setup where third-person perspectives are captured by calibrated Intel RealSense D435i cameras. 
We collect training demonstration data from five different camera perspectives, which provides sufficient viewpoint diversity for VLA learning while avoiding the excessive setup complexity of denser arrays to facilitate easy real-world deployment. 
For evaluation, we employ three testing cameras (Cam 2, Cam 3, and Cam 4). 
To evaluate viewpoint robustness, we rotate each testing camera by $5^\circ$, $10^\circ$, and $15^\circ$ around the robot base from its default position ($0^\circ$). 
We limit the viewpoint offsets up to $15^\circ$ to cover typical camera perturbations (e.g., sensor drifts) in daily operations, beyond which both the baseline and our model perform poorly, rendering further evaluation impractical. 
We evaluate on five common household tasks: \textit{put the basket upright}, \textit{pick up the banana and place to the circular basket}, \textit{push the cabbage near the pineapple}, \textit{wipe the table with the cloth}, and \textit{pick up the cup and place it inside the bowl}. 
For each task, we collect 100 demonstrations per training viewpoint. 
During evaluation, we run 20 episodes per task and viewpoint offset on each testing camera. 

\begin{table}[t]
\centering
\caption{
\textbf{Generalization to repositioned cameras on real-world robot experiments.}
Success rates (\%) under calibration-free viewpoint shifts, relative to the canonical view ($0^\circ$).}
\label{tab:real_world_robustness}
\scriptsize
\setlength{\tabcolsep}{3pt}
\begin{tabularx}{\columnwidth}{ll*{8}{>{\centering\arraybackslash}X}}
\toprule
\multirow{2}{*}{Model} & \multirow{2}{*}{Task} & \multicolumn{2}{c}{0$^\circ$ (Canonical)} & \multicolumn{2}{c}{5$^\circ$} & \multicolumn{2}{c}{10$^\circ$} & \multicolumn{2}{c}{15$^\circ$} \\
\cmidrule(lr){3-4} \cmidrule(lr){5-6} \cmidrule(lr){7-8} \cmidrule(lr){9-10}
& & Baseline & Ours & Baseline & Ours & Baseline & Ours & Baseline & Ours \\
\midrule
\multirow{5}{*}{$\pi_0$~\cite{pi0}} & Basket Upright & 75.0 & 88.3 & 58.3 & 80.0 & 31.7 & 68.3 & 16.7 & 48.3 \\
& Pick \& Place Banana & 53.3 & 80.0 & 45.0 & 68.3 & 33.3 & 50.0 & 0.0 & 18.3 \\
& Push Cabbage & 45.0 & 81.7 & 50.0 & 70.0 & 33.3 & 43.3 & 16.7 & 31.7 \\
& Wipe Table & 75.0 & 68.3 & 51.7 & 71.7 & 41.7 & 51.7 & 23.3 & 18.3 \\
& Pick \& Place Cup & 68.3 & 76.7 & 61.7 & 50.0 & 56.7 & 63.3 & 23.3 & 30.0 \\
\midrule
& Mean & 63.3 & 79.0 & 53.3 & 68.0 & 39.3 & 55.3 & 16.0 & 29.3 \\
\midrule
\multirow{5}{*}{GR00T N1.7~\cite{gr00t}} & Basket Upright & 95.0 & 96.7 & 88.3 & 100.0 & 36.7 & 73.3 & 0.0 & 45.0 \\
& Pick \& Place Banana & 48.3 & 70.0 & 28.3 & 53.3 & 13.3 & 30.0 & 0.0 & 8.3 \\
& Push Cabbage & 65.0 & 73.3 & 53.3 & 81.7 & 65.0 & 66.7 & 30.0 & 50.0 \\
& Wipe Table & 68.3 & 86.7 & 56.7 & 63.3 & 35.0 & 48.3 & 26.7 & 36.7 \\
& Pick \& Place Cup & 46.7 & 76.7 & 33.3 & 63.3 & 28.3 & 46.7 & 16.7 & 25.0 \\
\midrule
& Mean & 64.7 & 80.7 & 52.0 & 72.3 & 35.7 & 53.0 & 14.7 & 33.0 \\
\bottomrule
\end{tabularx}
\VspaceL
\end{table}

\noindent \textbf{Generalization to Repositioned Cameras.}
Table~\ref{tab:real_world_robustness} reports the results of our real-world robot experiments, evaluating the generalization performance of CamVLA under repositioned cameras with calibration-free viewpoint shifts. 
The reported success rates are averaged across all three cameras under the corresponding offset. 
We observe that under the canonical viewpoint ($0^\circ$), CamVLA substantially outperforms the standard baselines, achieving average success rates of 79.0\% (vs. 63.3\% for $\pi_0$) and 80.7\% (vs. 64.7\% for GR00T N1.7). 
Under the extreme $15^\circ$ offset, while the baseline success rates collapse to 16.0\% ($\pi_0$) and 14.7\% (GR00T), CamVLA maintains robust performance at 29.3\% and 33.0\%, respectively, demonstrating exceptional zero-shot generalization to uncalibrated viewpoint shifts. 
This is enabled by our CamVLA framework, which predicts camera-centric actions and utilizes the auxiliary Geometric Head to dynamically calibrate action outputs from monocular inputs during inference, highlighting the practicality of CamVLA in unstructured settings where camera positions may shift. 

To evaluate the accuracy of the Geometric Head in physical environments, we report the hand-eye errors across viewpoints in Table~\ref{tab:real_world_extrinsic_error}. 
We measure these errors against the ground-truth extrinsics from hand-eye calibration, showing that CamVLA maintains stable geometric grounding even under rotated test viewpoints. 
Despite the large errors at $15^\circ$, CamVLA outperforms the baseline because our relative-action formulation physically isolates translation errors, while the rotation error (under 10$^\circ$) remains within the tolerance of the closed-loop policy (as detailed in Sec.\,B of Supp.). 

\noindent \textbf{Computational Efficiency.} We report the parameter count, FLOPs, and inference time of CamVLA in Table~\ref{tab:efficiency}. 
To evaluate its real-world deployment feasibility, we measure the inference speed on a single consumer-grade NVIDIA RTX 4090 GPU. 
As shown, our auxiliary Geometric Head introduces only an extra 6.30~M parameters (0.19\% overhead), 1.0~G FLOPs (0.15\% overhead), and increases inference latency by only 1~ms (62~ms vs. 61~ms). 
With a 20-step trajectory executed at 10~Hz, the 62~ms inference easily runs in parallel with robot motion, presenting no bottleneck for real-time deployment. 

\begin{table}[t]
\centering
\begin{minipage}{0.52\textwidth}
\centering
\caption{\textbf{Real-world hand-eye matrix estimation errors.} 
Translation and rotation errors under different viewpoint offsets.}
\label{tab:real_world_extrinsic_error}
\vspace{4pt}
\scriptsize
\setlength{\tabcolsep}{2.5pt}
\begin{tabularx}{\linewidth}{l*{4}{>{\centering\arraybackslash}X}}
\toprule
Metric & 0$^\circ$ & 5$^\circ$ & 10$^\circ$ & 15$^\circ$ \\
\midrule
Trans. (cm) & 1.35 &2.12 & 7.91 & 27.16 \\
Rot. ($^\circ$) & 2.49 &4.73 & 5.98 & 9.39 \\
\bottomrule
\end{tabularx}
\end{minipage}
\hfill
\begin{minipage}{0.45\textwidth}
\centering
\caption{\textbf{Computational Efficiency.} 
Comparison of parameter count, FLOPs, and inference time on a single GeForce RTX 4090.}
\label{tab:efficiency}
\vspace{4pt}
\scriptsize
\setlength{\tabcolsep}{3pt}
\begin{tabularx}{\linewidth}{l*{3}{>{\centering\arraybackslash}X}}
\toprule
Model & Params (M) &FLOPs (G) & Inference (ms) \\
\midrule
$\pi_0$ \cite{pi0} &3238.1 &660.9 &61 \\
CamVLA &3244.4 &661.9 &62 \\
\bottomrule
\end{tabularx}
\end{minipage}
\VspaceL
\end{table} 

\subsection{Ablation Studies}
We ablate key design choices of CamVLA on RLBench~\cite{james2020rlbench} with $\pi_0$~\cite{pi0} as the baseline model, reporting the average success rate across unseen viewpoints for all evaluated tasks. 

\begin{wraptable}{r}{0.5\textwidth}
\centering
\vspace{-12pt}
\caption{\textbf{Generalization and hand-eye error across training distributions.} Success rates (\%) and pose estimation errors under different training intervals. CamVLA$^\dagger$ uses ground-truth extrinsics.}
\label{tab:training_range}
\scriptsize
\setlength{\tabcolsep}{2.5pt}
\begin{tabularx}{\linewidth}{c|*3{>{\centering\arraybackslash}X}|cc}
\toprule
\multirow{2}{*}{Interval} & \multicolumn{3}{c|}{Success Rate (\%)} & \multicolumn{2}{c}{Estimation Error} \\
& $\pi_0$ & CamVLA & CamVLA$^\dagger$ & Trans. (cm) & Rot. ($^\circ$) \\
\midrule
15$^\circ$ & 33.2 & 51.4 & 52.3 & 4.69 & 1.41 \\
30$^\circ$ & 25.5 & 34.0 & 40.0 & 19.71 & 4.77 \\
45$^\circ$ & 16.8 & 21.2 & 26.3 & 34.83 & 8.28 \\
\bottomrule
\end{tabularx}
\vspace{-9pt}
\end{wraptable}
\noindent \textbf{Density of Training Viewpoints.} 
We examine how the density of camera viewpoints in the training data affects generalization. 
We compare three training configurations with viewpoint sampling intervals of 15$^\circ$, 30$^\circ$, and 45$^\circ$. 
As shown in Table~\ref{tab:training_range}, denser viewpoint sampling leads to improved success rates under novel perspectives. 
As training viewpoints become sparser (30$^\circ$ and 45$^\circ$ intervals), the performance of all configurations inevitably degrades. 
Nevertheless, CamVLA consistently maintains a much higher unseen success rate (e.g., 34.0\% under 30$^\circ$ compared to 25.5\% of the baseline). 
The small performance gap between our method and GT under the $15^\circ$ interval (51.4\% vs. 52.3\%) confirms the high accuracy of our self-predicted extrinsics ($1.41^\circ$ rotation error). 
Even under the extremely sparse $45^\circ$ configuration, where self-localization errors increase ($8.28^\circ$), CamVLA still achieves a 21.2\% success rate, which is close to the GT performance of 26.3\% and outperforms the baseline by 4.4\%. 
These results demonstrate that our method is highly robust to self-localization errors. 
Notably, even the configuration with ground-truth extrinsics (CamVLA$^\dagger$) suffers similar degradation, confirming that this decline is primarily driven by visual representation shifts under unseen perspectives rather than hand-eye regression errors. 
Furthermore, while CamVLA operates without calibration, our framework is compatible with incorporating online hand-eye calibration systems to replace predicted extrinsics for better performance. 

\noindent \textbf{Viewpoint Intervals.}
To provide a more granular understanding of the local generalization capabilities, we present the detailed relative drop for each individual unseen viewpoint interval in Table~\ref{tab:detailed_drop_intervals}.
We define the \textit{Relative Drop (\%)} as $\text{Drop} = (\max(S_{\theta_1}, S_{\theta_2}) - S_{\text{unseen}}) / \max(S_{\theta_1}, S_{\theta_2})$, where $S_{\theta_1}$ and $S_{\theta_2}$ denote the success rates at the two adjacent training viewpoints, and $S_{\text{unseen}}$ represents the success rate at the unseen viewpoint. 
CamVLA achieves an average relative drop of 4.1\%, which outperforms the baseline (8.9\%), showing exceptional viewpoint invariance. 
In the viewpoint interval of $[30^\circ, 45^\circ]$, where the baseline suffers a severe performance drop of 20.8\%, ours maintains an exceptionally stable drop of only 0.9\%, showing robust invariance to viewpoint shifts. 

\noindent \textbf{State and Action Representations.}
As shown in Table~\ref{tab:calibration_free}, we evaluate the impact of different action spaces, proprioceptive states, and calibration requirements.
Shifting the action output from the robot base frame to the camera frame yields a significant performance gain, boosting the success rate from 33.2\% ($\pi_0$) to 51.4\% (CamVLA).
Using self-predicted instead of ground-truth extrinsics leads to only a marginal performance drop (e.g., 0.9\% gap between CamVLA$^\dagger$ and CamVLA, and 0.2\% gap between CamVLA$^{\ddagger}$ and CamVLA$^*$), validating the high accuracy of our auxiliary Geometric Head.
Although utilizing the base-frame proprioceptive state instead of the camera-frame state (comparing CamVLA with CamVLA$^*$) incurs a 0.3\% performance drop, it represents a key design choice that completely eliminates the need for camera calibration in the state observation phase, enabling true calibration-free deployment. 

\begin{table}[t]
\centering
\caption{\textbf{Relative performance drop across individual viewpoint intervals.} We compare the unseen success rates and their relative drops from adjacent seen viewpoints across unseen intervals.}
\label{tab:detailed_drop_intervals}
\scriptsize
\setlength{\tabcolsep}{2.5pt}
\begin{tabularx}{\linewidth}{c|*3{>{\centering\arraybackslash}X}|*3{>{\centering\arraybackslash}X}}
\toprule
\multirow{2}{*}{Viewpoint Interval} & \multicolumn{3}{c|}{$\pi_0$} & \multicolumn{3}{c}{CamVLA} \\
& Seen (\%) & Unseen (\%) & Drop (\%) & Seen (\%) & Unseen (\%) & Drop (\%) \\
\midrule
$[0^\circ, 15^\circ]$  & 42.0 & 32.8 & 21.8 & 70.0 & 58.7 & 16.2 \\
$[15^\circ, 30^\circ]$ & 42.0 & 44.0 & -4.8  & 70.0 & 61.5 & 12.1 \\
$[30^\circ, 45^\circ]$ & 37.7 & 29.8 & 20.8 & 57.7 & 57.2 & 0.9  \\
$[45^\circ, 60^\circ]$ & 35.7 & 37.0 & -3.7  & 50.0 & 51.5 & -3.0  \\
$[60^\circ, 75^\circ]$ & 35.0 & 30.0 & 14.3 & 43.0 & 43.8 & -1.9  \\
$[75^\circ, 90^\circ]$ & 26.7 & 25.3 & 5.0  & 35.7 & 35.7 & 0.0   \\
\midrule
Average & 36.5 & 33.2 & 8.9 & 54.4 & 51.4 & 4.1 \\
\bottomrule
\end{tabularx}
\VspaceL
\end{table} 

\begin{table}[t]
\centering
\caption{\textbf{State and action space ablations.} 
Comparison of various proprioceptive state representations, action outputs, and calibration requirements.}
\label{tab:calibration_free}
\scriptsize
\setlength{\tabcolsep}{4pt}
\begin{tabular*}{\columnwidth}{@{\extracolsep{\fill}}lccccc}
\toprule
Model Variant & State Input & Action Output & Hand-Eye Source & Calibration-Free? & Success Rate (\%) \\
\midrule
$\pi_0$ & Base & Base & None & \ding{51} & 33.2 \\
\midrule
CamVLA$^{\ddagger}$ & Camera & Camera & Ground-Truth & \ding{55} & 51.9 \\
CamVLA$^*$ & Camera & Camera & Self-Predicted & \ding{55} & 51.7 \\
\midrule

CamVLA$^\dagger$ & Base & Camera & Ground-Truth & \ding{55} & 52.3 \\
CamVLA & Base & Camera & Self-Predicted & \ding{51} & 51.4 \\
\bottomrule
\end{tabular*}
\raggedright
\VspaceL
\end{table}

\section{Conclusion and Limitations}
In this work, we present CamVLA, a calibration-free VLA model that achieves viewpoint robustness without external camera information. 
By decoupling the policy into an Action Head that predicts the camera-centric action and a Geometric Head that regresses the hand-eye matrix from a single monocular RGB image, CamVLA isolates camera-centric action generation from camera-perspective geometric grounding, composing their outputs into base-frame actions through a deterministic transformation. 
Experiments in simulation and on real hardware demonstrate substantial gains over strong VLA baselines under unseen camera configurations. 
By replacing given geometry with learned geometry, CamVLA brings VLA policies one step closer to robust and calibration-free manipulation under unseen camera viewpoints in unstructured environments. 

While CamVLA achieves viewpoint robustness for the third-person perspective, it only utilizes a single third-person camera and does not consider viewpoint perturbations of the wrist-mounted camera in multi-camera systems. Furthermore, although robust to typical shifts, the model struggles under extreme viewpoint changes and high-precision tasks due to out-of-distribution visual features and hand-eye regression errors. In future work, we plan to improve geometric regression robustness under large viewpoint shifts and investigate wrist-mounted camera perturbations. 

\bibliography{ref}

\clearpage
\appendix
{\noindent\Large\textbf{Supplementary Material}}
\newline

This supplementary material covers the following details:
\begin{itemize}
    \item Additional Implementation details (Sec.~\ref{sec:implementation_details}). 
    \item Additional Mathematical derivation (Sec.~\ref{sec:translation_independence}). 
    \item Additional ablation studies (Sec.~\ref{sec:ablation_feature}). 
    \item Additional visualization results (Sec.~\ref{sec:visualization}). 
\end{itemize}

\section{Additional Implementation Details}
\label{sec:implementation_details}

\noindent \textbf{Architecture and Training.} 
CamVLA is trained in a multi-task manner on 8 NVIDIA H100 80GB GPUs, and we subsequently evaluate its performance across different tasks. 
We instantiate CamVLA upon foundational VLA architectures, specifically $\pi_0$~\cite{pi0} and GR00T N1.7~\cite{gr00t}, adhering to their respective training recipes and configurations. 
Visual inputs are restricted to single third-person monocular RGB images resized to $224 \times 224$ pixels. 
To evaluate robustness to third-person viewpoint changes and prevent the model from relying on the unaffected wrist view to bypass them, we exclude wrist cameras, which also simplifies hardware routing across robot joints. 
The predicted robot actions are parameterized as delta 6-DoF end-effector poses relative to the current state, where the rotation component is represented as a 3D axis-angle vector. 
The auxiliary Geometric Head is implemented as a lightweight three-layer Multi-Layer Perceptron (MLP) with GELU activations and a hidden dimension of 1024. 
It operates on the high-level semantic features extracted from the visual tokens of the backbone, which are aggregated via mean pooling before being fed into the MLP. 

\noindent \textbf{Simulation and Real-World Setup.} 
The simulation is powered by the CoppeliaSim engine, with expert demonstrations generated by the Open Motion Planning Library (OMPL)~\cite{sucan2012open}. 
In simulation, the control frequency is 20\,Hz and the model executes the first 5 steps of the predicted action trajectory per inference. 
For real-world experiments, raw demonstrations are collected at 30\,Hz, while training and deployment are conducted at 10\,Hz to ensure stable hardware response, with the model executing the first 20 steps of the predicted trajectory per inference. 
In both settings, ground-truth hand-eye matrices are used only as training supervision, and the model receives no extrinsic information at deployment. 

\noindent \textbf{Optimization.} 
We optimize the model end-to-end with a joint objective $\mathcal{L} = \mathcal{L}_{act} + \lambda \mathcal{L}_{ext}$, where $\mathcal{L}_{act}$ inherits the action prediction loss of the underlying baseline (e.g., flow-matching for $\pi_0$~\cite{pi0}) and the geometric grounding loss is the mean squared error $\mathcal{L}_{ext} = \sum_t \big(\|\tau_t - \hat{\tau}_t\|_2^2 + \|\omega_t - \hat{\omega}_t\|_2^2\big)$, with $\hat{\tau}_t, \hat{\omega}_t$ denoting the ground-truth translation and axis-angle rotation vectors. 
We set $\lambda = 0.1$. 

\section{Additional Mathematical Derivation}
\label{sec:translation_independence}

We provide the derivation showing why the execution of camera-centric delta actions is independent of the hand-eye translation vector.
Let the hand-eye transform $T_t \in SE(3)$ from the camera frame to the robot base frame be parameterized by a translation vector $\tau_t \in \mathbb{R}^3$ and an axis-angle rotation vector $\omega_t \in \mathbb{R}^3$.
We convert $\omega_t$ to a rotation matrix $R_t \in SO(3)$ (e.g., via Rodrigues' rotation formula) to construct the homogeneous transformation matrix:
\begin{equation}
    T_t =
    \begin{bmatrix}
        R_t & \tau_t \\
        0 & 1
    \end{bmatrix},
    \qquad R_t \in SO(3), \ \tau_t \in \mathbb{R}^3 .
\end{equation}
For two end-effector positions expressed in the camera frame, $p_{c,t}^{(0)}$ and $p_{c,t}^{(1)}$ (denoting the initial and target positions of a delta action, respectively), their corresponding base-frame positions are
\begin{equation}
    p_{b,t}^{(i)} = R_t p_{c,t}^{(i)} + \tau_t, \qquad i \in \{0,1\}.
\end{equation}
The base-frame relative translation is therefore
\begin{align}
    \Delta p_{b,t}
    &= p_{b,t}^{(1)} - p_{b,t}^{(0)} \nonumber \\
    &= \left(R_t p_{c,t}^{(1)} + \tau_t\right)
     - \left(R_t p_{c,t}^{(0)} + \tau_t\right) \nonumber \\
    &= R_t \left(p_{c,t}^{(1)} - p_{c,t}^{(0)}\right)
     = R_t \Delta p_{c,t}.
\end{align}
Thus, $\tau_t$ cancels exactly for relative translations.
Delta rotations are also independent of the hand-eye translation vector.
Let $Q_{c,t}^{(0)}, Q_{c,t}^{(1)} \in SO(3)$ denote the two end-effector orientation matrices in the camera frame, with $Q_{b,t}^{(i)} = R_t Q_{c,t}^{(i)}$ in the base frame.
The relative rotation in the base frame satisfies
\begin{align}
    \Delta Q_{b,t}
    &= Q_{b,t}^{(1)} \left(Q_{b,t}^{(0)}\right)^\top \nonumber \\
    &= R_t Q_{c,t}^{(1)} \left(Q_{c,t}^{(0)}\right)^\top R_t^\top
     = R_t \Delta Q_{c,t} R_t^\top .
\end{align}
Using the equivariance of the matrix logarithm under rotation conjugation,
\begin{equation}
    [\Delta r_{b,t}]_\times
    = \log(\Delta Q_{b,t})
    = R_t \log(\Delta Q_{c,t}) R_t^\top
    = [R_t \Delta r_{c,t}]_\times ,
\end{equation}
which gives $\Delta r_{b,t} = R_t \Delta r_{c,t}$.
Consequently, both components of the executed delta action depend only on $R_t$:
\begin{equation}
    \Delta A_{b,t} =
    [R_t \Delta p_{c,t}, \ R_t \Delta r_{c,t}, \ g_t]. 
\end{equation}

\section{Additional Ablation Studies}
\label{sec:ablation_feature}

\begin{figure}[t]
  \centering
  \includegraphics[width=0.55\textwidth]{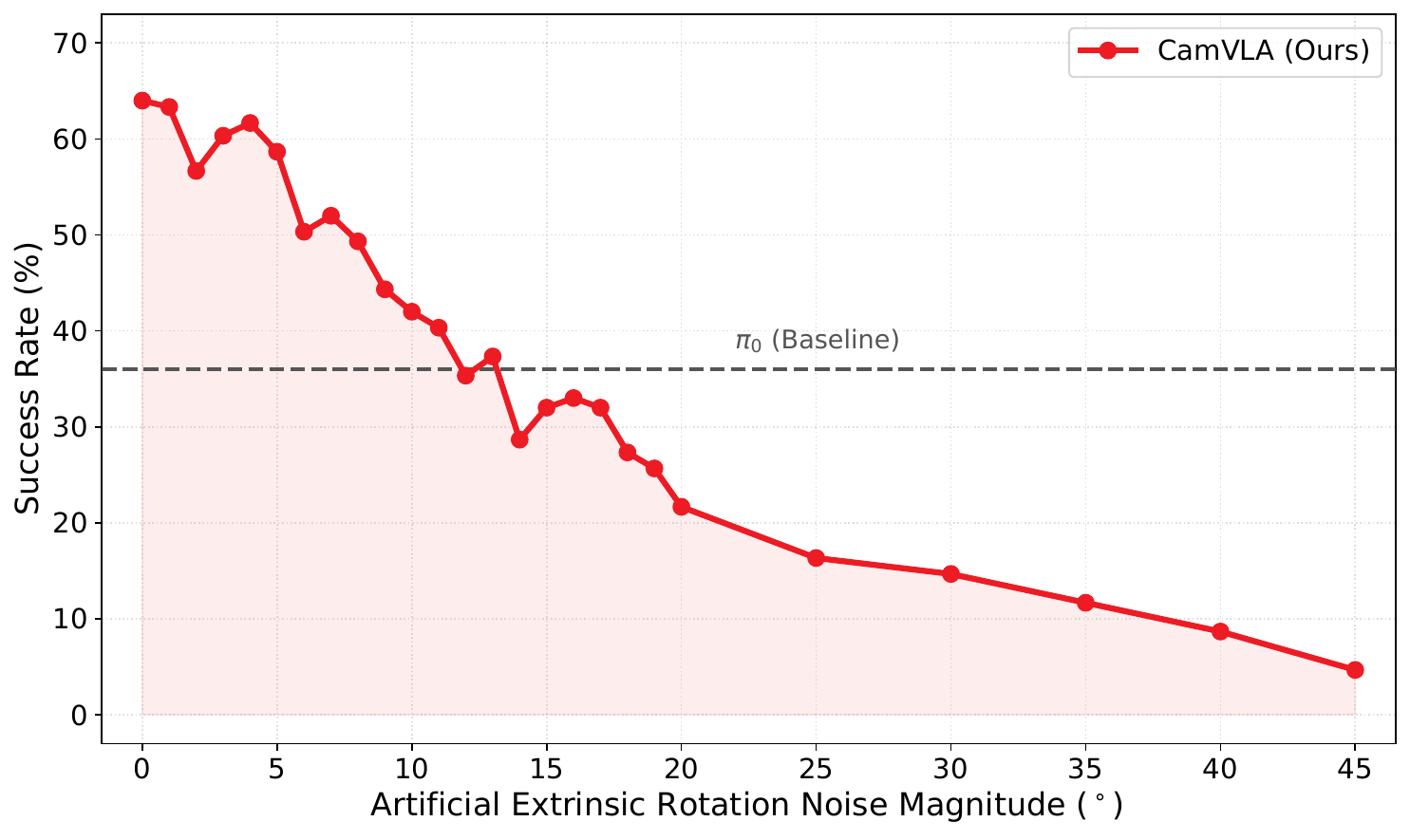}
  \caption{\textbf{Robustness under artificial extrinsic rotation noise.} Success rate (\%) under varying levels of random rotation noise (ranging from $0^\circ$ to $45^\circ$) applied to the ground-truth hand-eye rotation matrix during execution.}
  \label{fig:supp_rotation_noise}
\end{figure}

\noindent \textbf{Extrinsic Noise Robustness.} 
To evaluate the robustness of CamVLA to hand-eye matrix estimation errors, we conduct an ablation study with artificial rotation noise during evaluation. 
At test time, we systematically inject random rotation noise into the ground-truth hand-eye rotation matrix across a wide range of noise levels, which consists of a $0^\circ$ reference, $1^\circ$--$20^\circ$ in $1^\circ$ increments, and $25^\circ$--$45^\circ$ in $5^\circ$ increments around a random 3D unit axis. 
To closely simulate the dynamic prediction fluctuations of the geometric head, the rotation noise is independently re-sampled at each planning step rather than applying a static bias across the entire episode.
We explicitly omit translation noise because, as derived above, delta action execution is independent of the hand-eye translation vector under our relative-action formulation. 
As shown in Figure~\ref{fig:supp_rotation_noise}, the policy is highly robust to small rotation perturbations: the success rate decreases only moderately from the 64.0\% no-noise reference to 63.3\% at $1^\circ$ and 58.7\% at $5^\circ$. 
Notably, even under injected rotation noise magnitudes of up to $12^\circ$, CamVLA still outperforms or remains highly comparable to the noise-free baseline $\pi_0$ (36.0\%). 
Crucially, our geometric head predicts hand-eye rotation with a mean error of less than $2^\circ$ (specifically $1.41^\circ$ as shown in the ablation table of the main text), a high-precision regime where CamVLA suffers almost no performance degradation. 
This experiment validates the feasibility of our calibration-free framework, demonstrating that while closed-loop VLA policies naturally possess an inherent tolerance to small execution deviations, our decoupled design (which parallelly predicts camera-centric actions and hand-eye poses) successfully extends this feedback robustness to encompass tolerance against hand-eye rotation errors as well, explaining why CamVLA achieves high success rates despite using self-predicted, imperfect extrinsics. 

\begin{wraptable}{r}{0.5\columnwidth}
\centering
\vspace{-0pt}
\caption{\textbf{Ablation on hand-eye pose representation.} 
Comparison of success rates and geometric errors between predicting full 6-DoF extrinsics and regressing rotation only.}
\label{tab:rotation_only}
\vspace{-6pt}
\scriptsize
\setlength{\tabcolsep}{1.5pt}
\begin{tabularx}{\linewidth}{l*{3}{>{\centering\arraybackslash}X}}
\toprule
Pose Representation & Success (\%) & Trans. (cm) & Rot. ($^\circ$) \\
\midrule
Full 6-DoF & 51.4 & 4.7 & 1.4 \\
Rotation Only & 51.3 & - & 1.6 \\
\bottomrule
\end{tabularx}
\vspace{-10pt}
\end{wraptable}

\noindent \textbf{Hand-Eye Pose Representation.}
We ablate the necessity of predicting the translation component of the hand-eye matrix. 
Under our relative-action formulation, policy execution depends only on the rotation component. 
To verify this, we compare the default 6-DoF configuration against a rotation-only variant. 
As shown in Table~\ref{tab:rotation_only}, predicting only rotation yields a 51.3\% success rate and a 1.6$^\circ$ rotation error, highly comparable to the default 6-DoF configuration (51.4\% success rate and 1.4$^\circ$).
This demonstrates that additionally predicting the translation component does not degrade task success or rotation estimation accuracy. 
Furthermore, regressing the full 6-DoF pose allows CamVLA to support potential absolute-action variants, enhancing framework versatility. 

\begin{wraptable}{r}{0.50\columnwidth}
\centering
\vspace{-12pt}
\caption{\textbf{Ablation on the visual feature source for hand-eye matrix prediction.} 
Comparison of success rates and geometric errors using different visual features. }
\label{tab:feature_source}
\vspace{-6pt}
\scriptsize
\setlength{\tabcolsep}{1.5pt}
\begin{tabularx}{\linewidth}{l*{3}{>{\centering\arraybackslash}X}}
\toprule
Feature Source & Success (\%) & Trans. (cm) & Rot. ($^\circ$) \\
\midrule
Image Encoder & 51.4 & 4.7 & 1.4 \\
Image Encoder (Detach) & 42.6 & 11.7 & 13.2 \\
\midrule
VLM Backbone & 53.5 & 14.7 & 3.7 \\
VLM Backbone (Detach) & 20.9 & 45.0 & 36.0 \\
\bottomrule
\end{tabularx}
\vspace{-10pt}
\end{wraptable}

\noindent \textbf{Visual Feature Sources.} 
We investigate how the visual feature source and gradient propagation affect hand-eye matrix prediction. 
As shown in Table~\ref{tab:feature_source}, using features from the Image Encoder yields the most precise geometric localization, with only 4.7~cm translation error and $1.4^\circ$ rotation error, and achieves a 51.4\% unseen success rate. 
Extracting features from the deeper VLM Backbone slightly improves the task success rate to 53.5\%, but substantially worsens geometric estimation (14.7~cm and $3.7^\circ$), indicating that high-level semantic features can support action generation while being less spatially precise for camera pose regression. 
Although the VLM Backbone configuration achieves a slightly higher success rate, we select the Image Encoder as our default configuration. 
The primary reason is that the VLM Backbone overfits to the discrete training viewpoints, producing geometric discontinuities and large localization spikes at unseen intermediate angles. 
 Such geometric instability poses significant risks during physical robot execution, where smooth and consistent hand-eye estimation is crucial for safety. 
By contrast, the Image Encoder configuration offers stable and continuous spatial grounding across the entire viewpoint spectrum, making it a more reliable choice for real-world deployment. 
Detaching gradients consistently hurts both localization and control. 
Specifically, Image Encoder detachment reduces success from 51.4\% to 42.6\%, while VLM Backbone detachment collapses the success rate to 20.9\% with severe pose errors (45.0~cm and $36.0^\circ$). 
These results suggest that the Geometric Head benefits from end-to-end feature adaptation, and we therefore use Image Encoder features without detachment as the default configuration for its superior spatial grounding stability. 

\section{Additional Visualization Results}
\label{sec:visualization}

\begin{figure*}[t]
  \centering
  \includegraphics[width=0.9\textwidth]{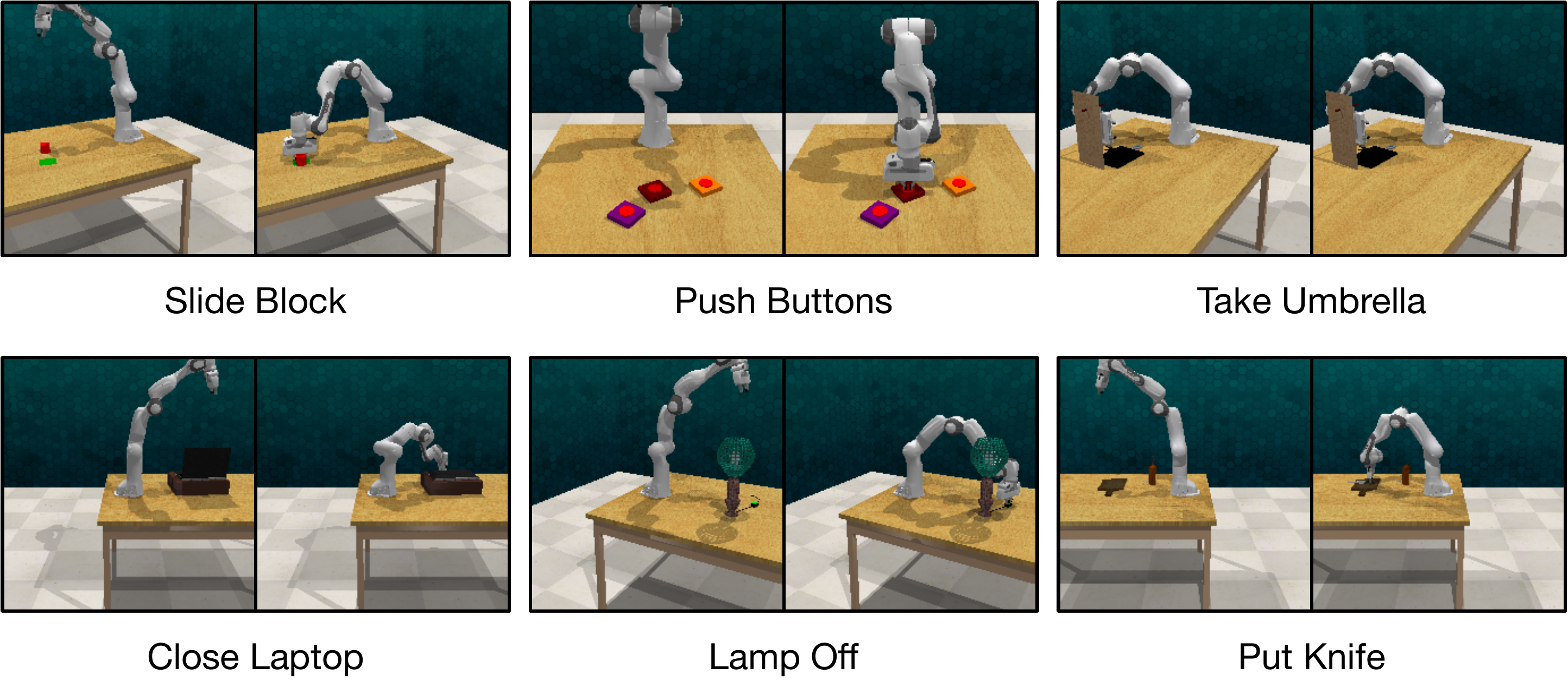}
  \caption{\textbf{Simulation tasks on RLBench benchmark.} 
We evaluate CamVLA across a diverse set of manipulation tasks, requiring both high-level semantic understanding and precise low-level control.}
  \label{fig:supp_sim_task}
\end{figure*}

\begin{figure}[t]
  \centering
  \includegraphics[width=\textwidth]{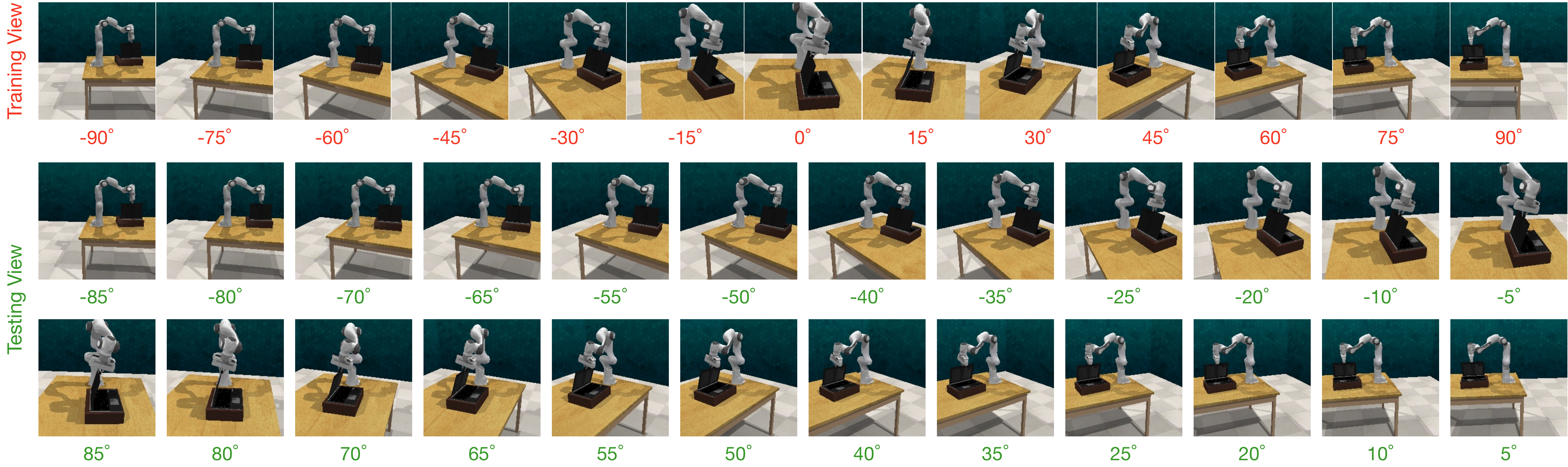}
  \caption{\textbf{Visualization of training and testing viewpoints on RLBench.} 
The training set consists of views sampled at 15$^\circ$ intervals (top row), while the testing set covers a dense range of unseen viewpoints (middle and bottom rows) to evaluate the zero-shot generalization capability of CamVLA.}
  \label{fig:sim_views}
\end{figure}

\begin{figure}[t]
  \centering
  \includegraphics[width=\textwidth]{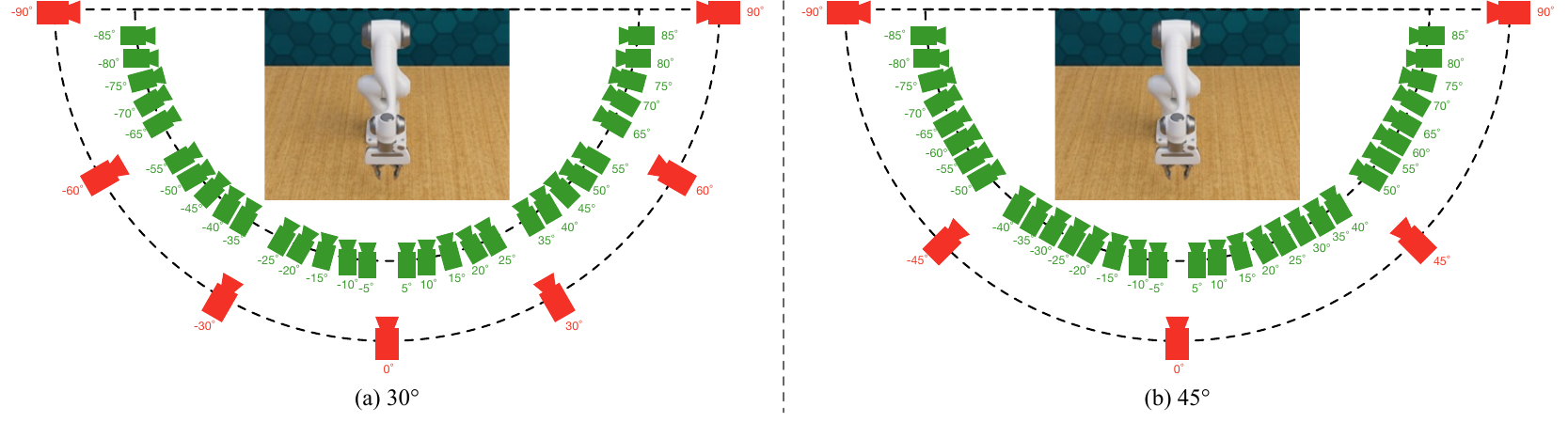}
  \caption{\textbf{Detailed range of camera viewpoints in simulation.} 
  (a) Training viewpoints sampled at 30$^\circ$ intervals and (b) training viewpoints sampled at 45$^\circ$ intervals. 
  Red and green cameras represent training and unseen testing viewpoints, respectively. 
  }
  \label{fig:supp_sim_camera}
\end{figure}

\begin{figure*}[t]
  \centering
  \includegraphics[width=1.0\textwidth]{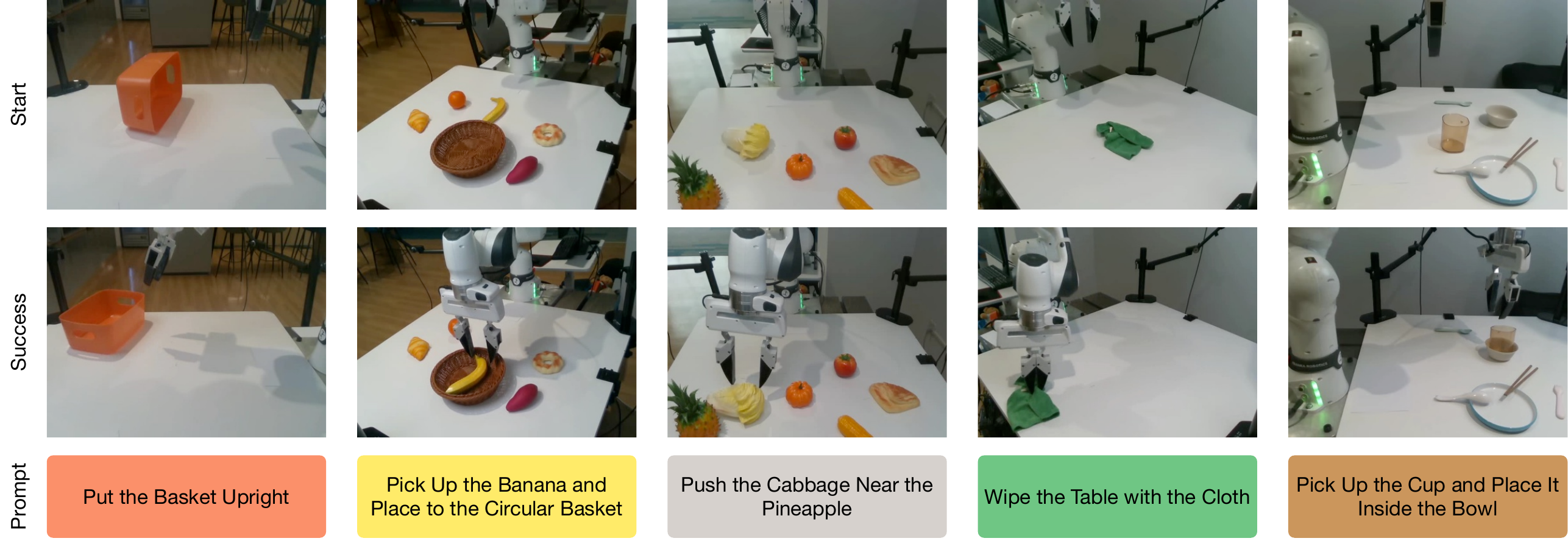}
  \caption{\textbf{Real-world evaluation tasks.} 
Five representative manipulation tasks involving diverse objects and interaction requirements.}
  \label{fig:supp_real_tasks}
\end{figure*}

\begin{figure}[t]
  \centering
  \includegraphics[width=\textwidth]{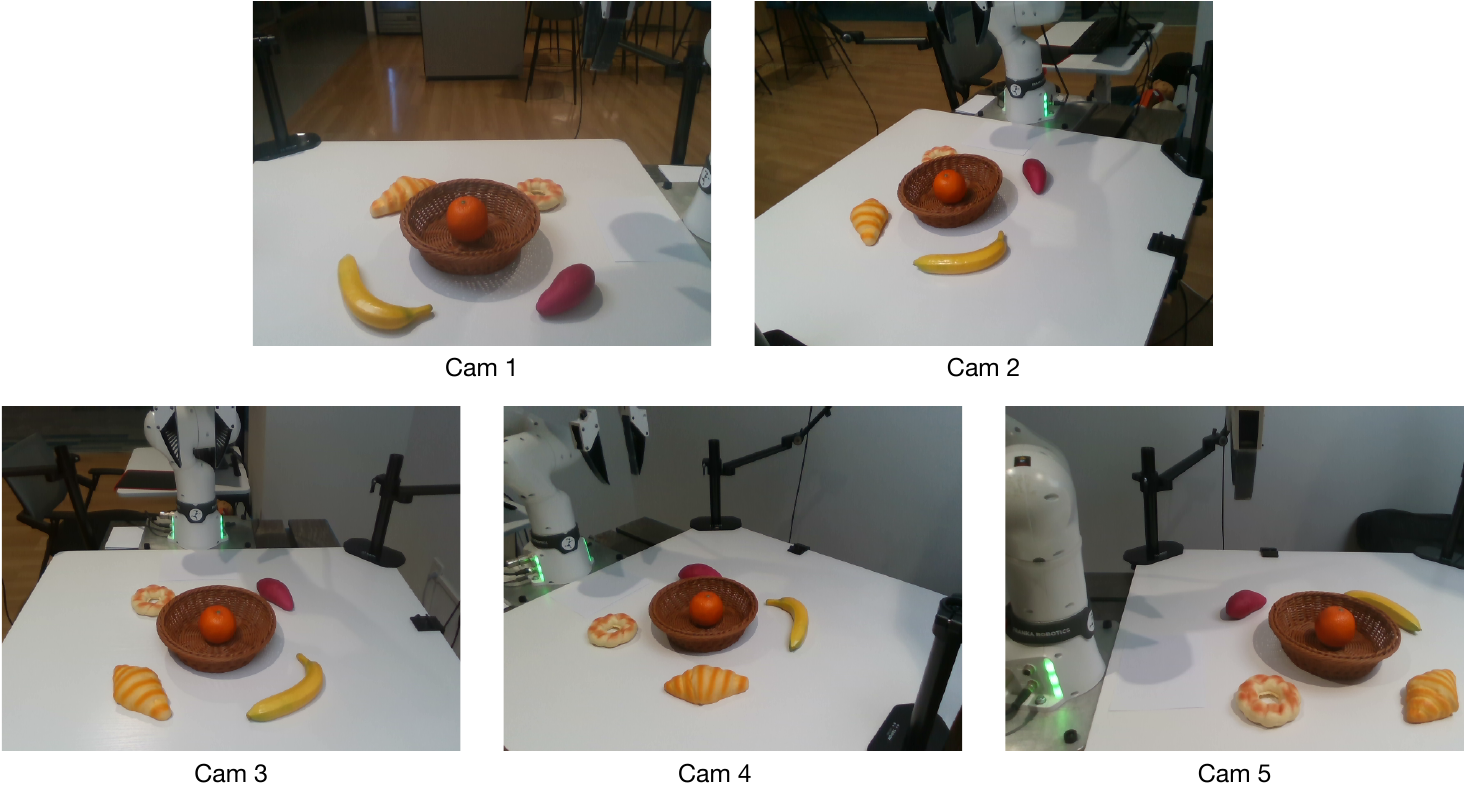}
  \caption{\textbf{Real-world training camera viewpoints.}
Five third-person training placements (Cam 1--Cam 5) used for demonstration collection.}
  \label{fig:supp_real_view}
\end{figure}

\begin{figure}[t]
  \centering
  \includegraphics[width=\textwidth]{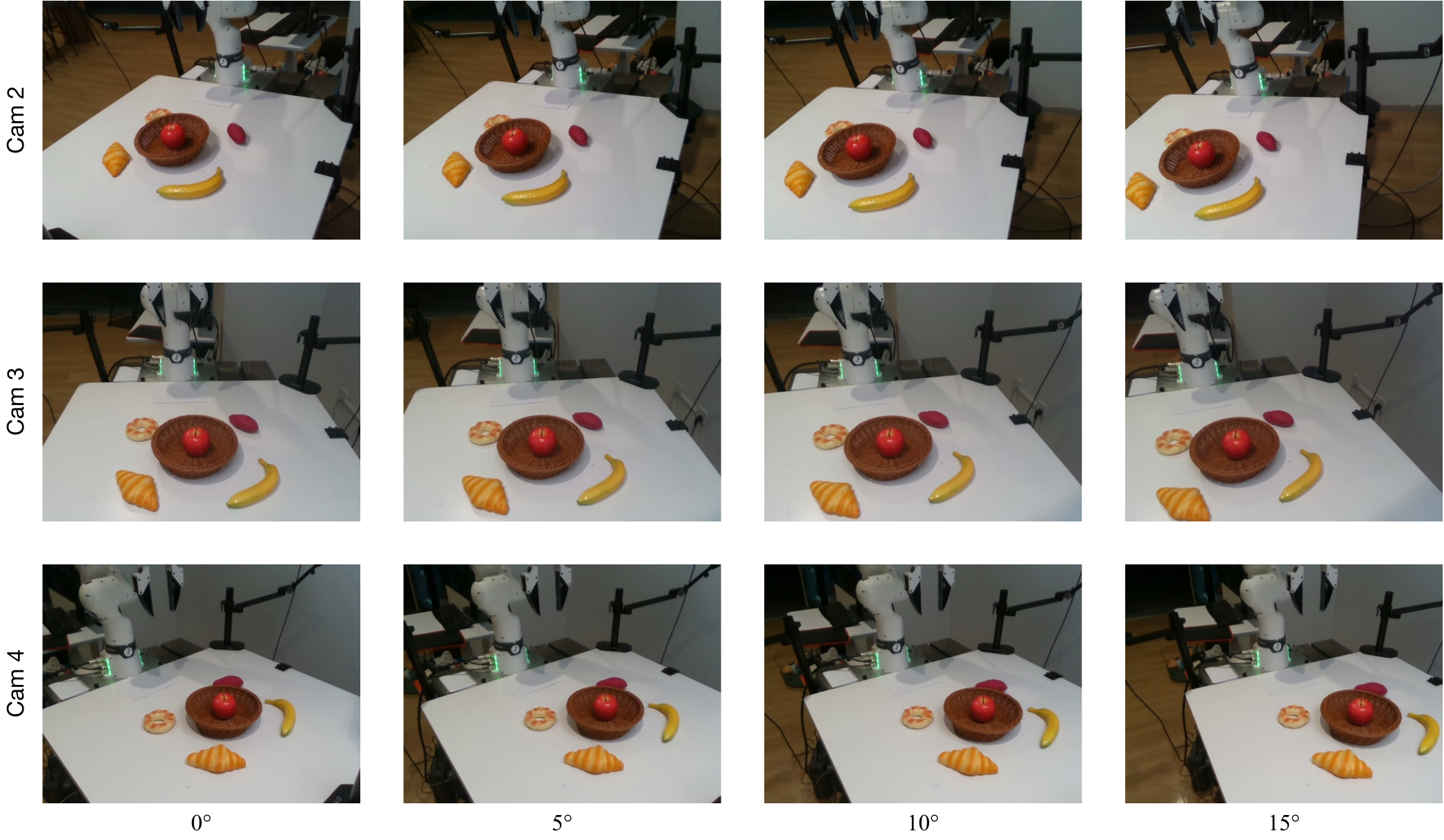}
  \caption{\textbf{Real-world unseen testing viewpoints.}
Cam 2, Cam 3, and Cam 4 are horizontally rotated by $5^\circ$, $10^\circ$, and $15^\circ$ from their canonical $0^\circ$ training position to form unseen testing views.}
  \label{fig:supp_real_camera}
\end{figure}

\noindent \textbf{Visualization of Setup and Tasks.}
Figure~\ref{fig:supp_sim_task} provides a visual overview of the 10 RLBench simulation tasks used in our evaluation. 
Figure~\ref{fig:sim_views} visualizes the visual range of both training and testing viewpoints in RLBench. 
Figure~\ref{fig:supp_sim_camera} provides a detailed illustration of the camera placement around the robot base for the training viewpoint distribution (ablation study shown in Table 5). 
Figure~\ref{fig:supp_real_tasks} illustrates the five representative household-style manipulation tasks used in our real-world evaluation.
Figure~\ref{fig:supp_real_view} illustrates the five training camera viewpoints (Cam 1--Cam 5) used for demonstration collection in our real-world hardware setup.
Figure~\ref{fig:supp_real_camera} further visualizes the unseen testing viewpoints, where Cam 2, Cam 3, and Cam 4 are perturbed by horizontal rotations of $5^\circ$, $10^\circ$, and $15^\circ$ relative to their canonical $0^\circ$ training position. 

\begin{figure*}[t]
  \centering
  \includegraphics[width=\textwidth]{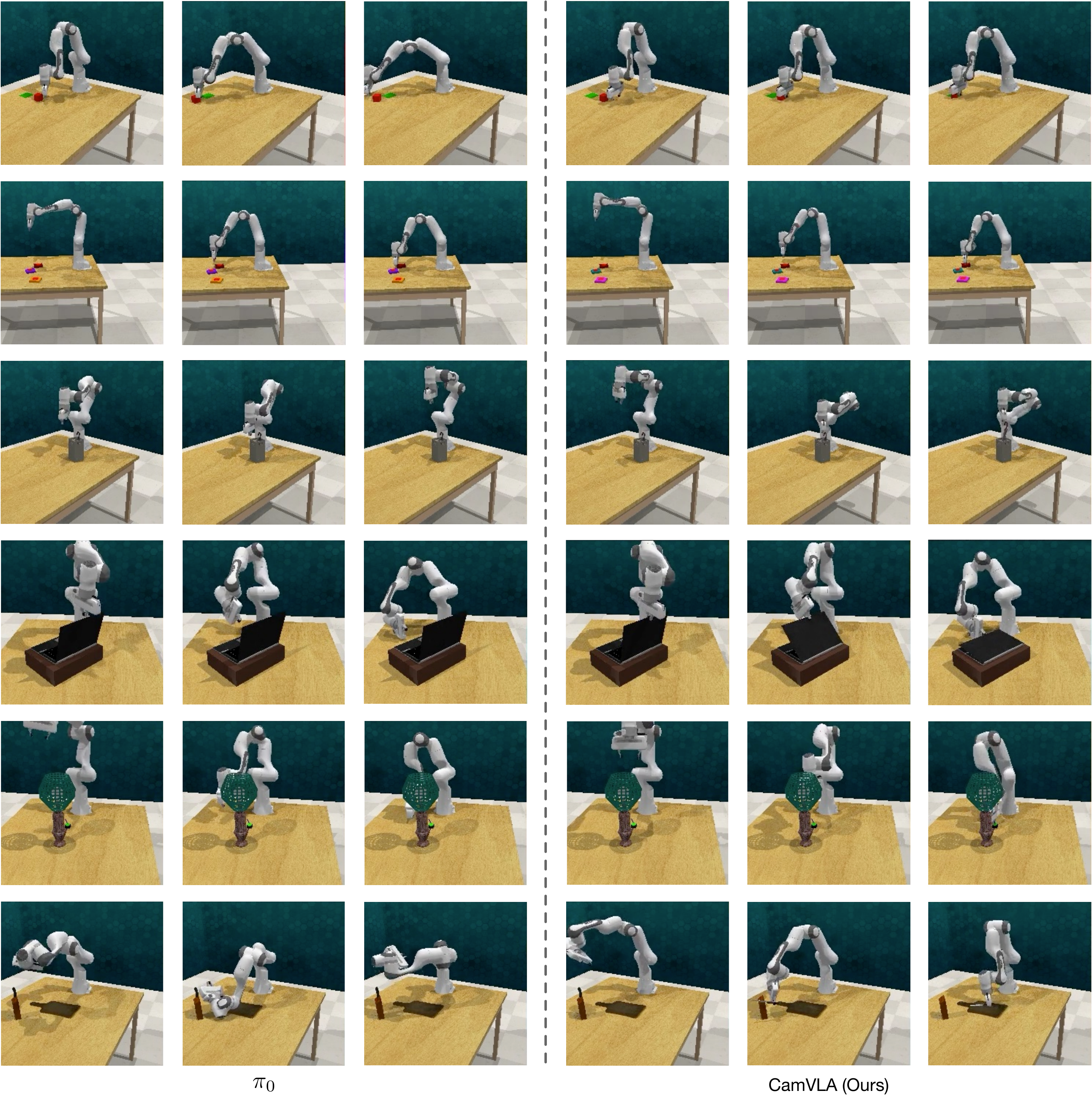}
  \caption{\textbf{Qualitative comparison between $\pi_0$ and CamVLA on RLBench under unseen cameras.}}
  \label{fig:supp_sim_compare}
\end{figure*}

\begin{figure*}[t]
  \centering
  \includegraphics[width=\textwidth]{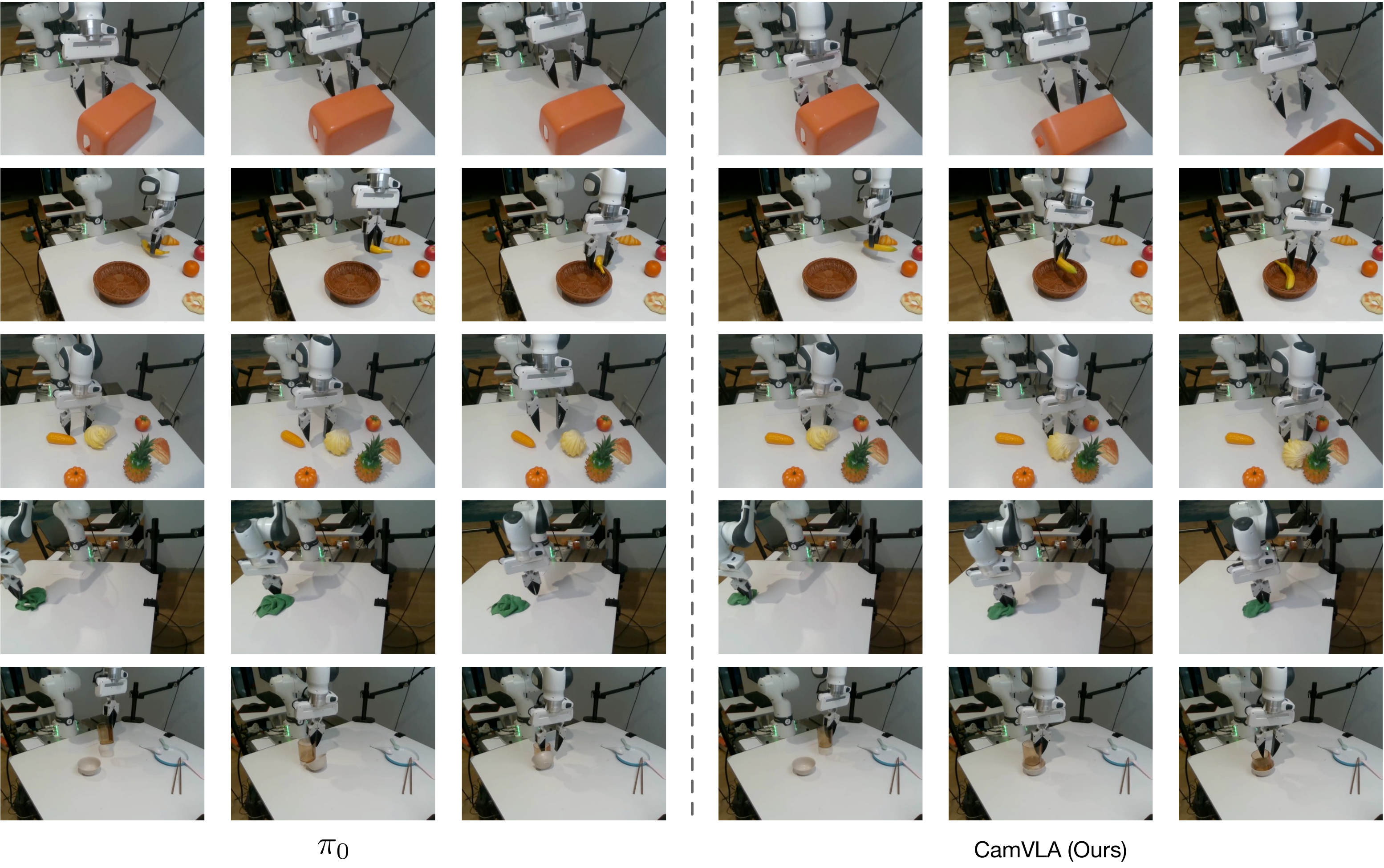}
  \caption{\textbf{Qualitative comparison between $\pi_0$ and CamVLA on real-world robot experiments under repositioned cameras.}}
  \label{fig:supp_real_compare}
\end{figure*}

\begin{figure*}[t]
  \centering
  \includegraphics[width=\textwidth]{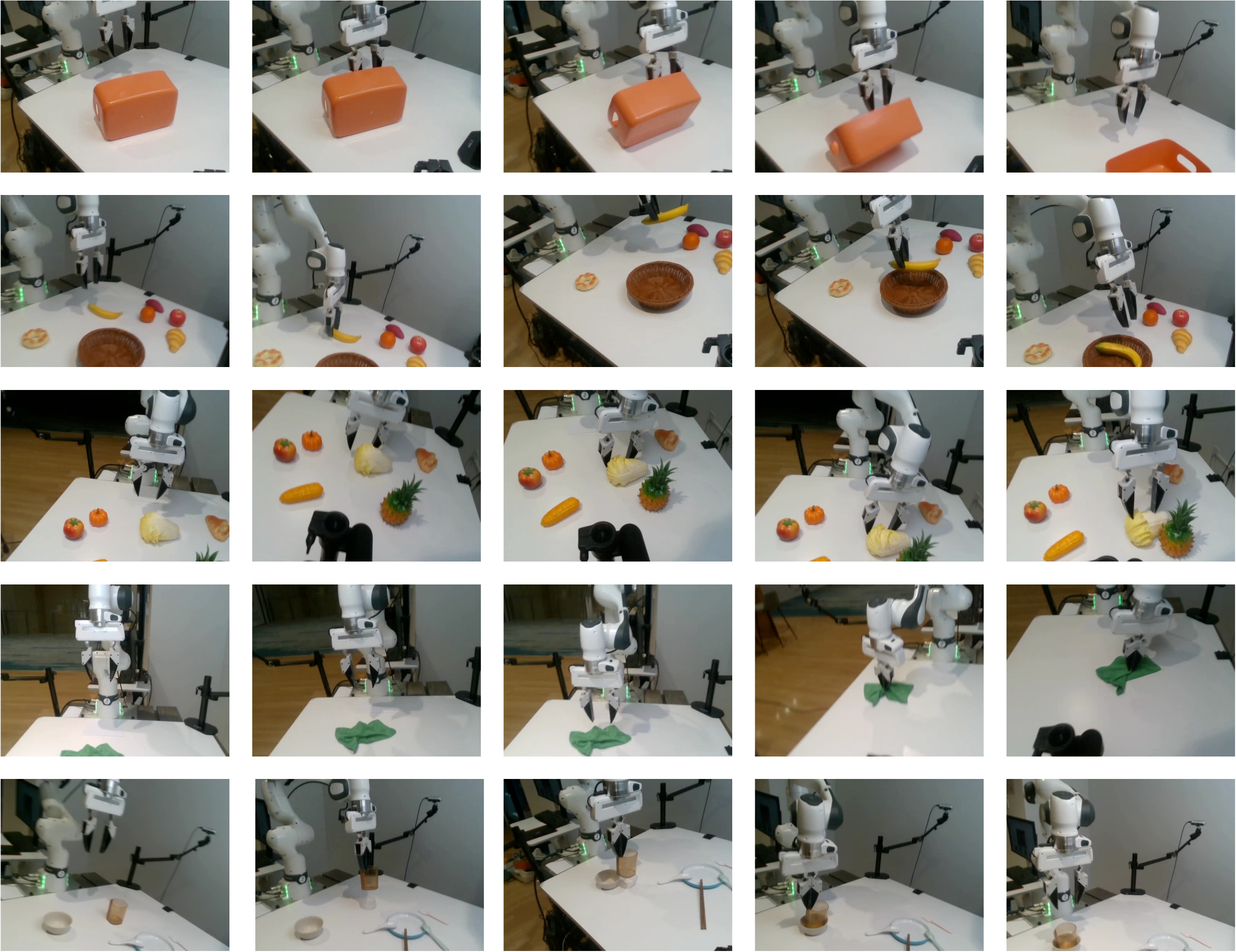}
  \caption{\textbf{Visualization of real-world experiments under dynamically hand-held moving cameras.} 
  We visualize the robot's execution and dynamic hand-eye pose tracking when the third-person camera is continuously moved by a human operator during deployment.}
  \label{fig:supp_moving}
\end{figure*}

\noindent \textbf{Qualitative Comparison.} 
Figure~\ref{fig:supp_sim_compare} and Figure~\ref{fig:supp_real_compare} provide qualitative comparisons of execution trajectories under unseen simulation viewpoints and repositioned real-world camera viewpoints, respectively. 
We compare the execution trajectories of the baseline $\pi_0$ and our CamVLA under unseen viewpoints. 
CamVLA maintains target-directed behavior and completes the tasks more robustly than the baseline $\pi_0$. 
Additionally, Figure~\ref{fig:supp_moving} visualizes the real-world deployment under continuously moving and hand-held cameras, echoing the drifting or hand-held camera scenarios and demonstrating the robust viewpoint generalization capability of CamVLA. 

\begin{figure*}[t]
  \centering
  \includegraphics[width=\textwidth]{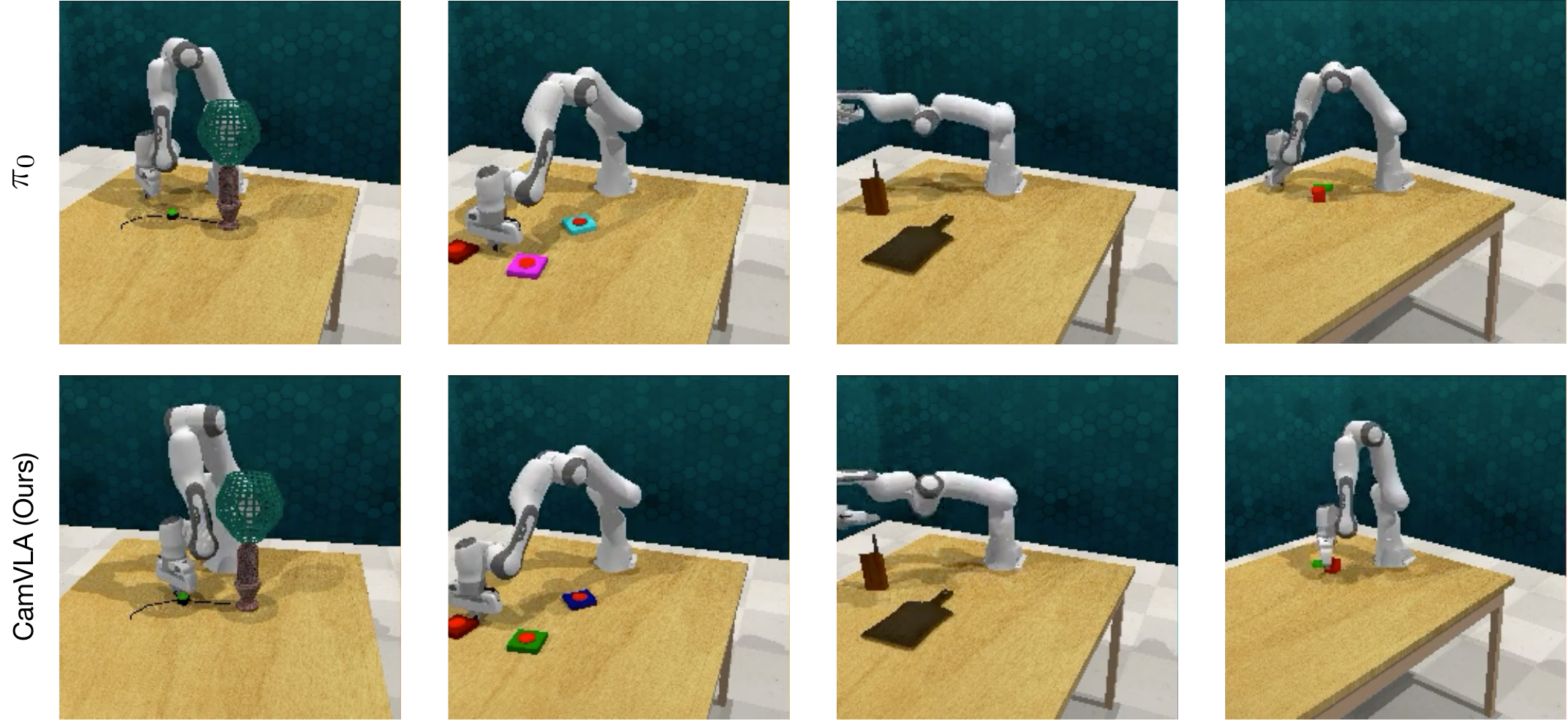}
  \caption{\textbf{Failure cases.} 
  We illustrate common failure modes such as boundary objects, actions exceeding the robot's physical workspace, and self-occlusion.}
  \label{fig:supp_fail}
\end{figure*}

\noindent \textbf{Failure Case Analysis.} 
Figure~\ref{fig:supp_fail} visualizes representative failure cases of both $\pi_0$ and CamVLA. 
These failures are primarily caused by target objects located at the boundary of the camera's field of view, actions exceeding the physical workspace of the robot arm, or visual self-occlusion. 
Notably, even in these failed cases, CamVLA's execution trajectories still exhibit more target-directed and reasonable behaviors compared to the baseline $\pi_0$. 

\newpage

\end{document}